\documentclass[runningheads]{llncs}
\usepackage{graphicx}
\usepackage{xcolor}
\usepackage{bm}
\usepackage{amsmath}
\usepackage{amssymb}
\usepackage{caption}
\usepackage{subcaption}
\usepackage{hyperref}

\usepackage[ruled,vlined]{algorithm2e}
\newcommand{\R}{\mathbb{R}}

\newcommand{\norm}[1]{\left\lVert#1\right\rVert}

\newcommand{\M}{\mathcal{M}}

\begin{document}
\title{Joint Geometric and Topological Analysis of Hierarchical Datasets}
\author{Lior Aloni\inst{1} \and
Omer Bobrowski\inst{1} \and
Ronen Talmon \inst{1}}

%\author{First Author \and
%Second Author \and
%Third Author}

%
\authorrunning{L. Aloni et al.}

\institute{Technion -- Israel Institute of Technology, Haifa, Israel \\
\email{\{lioral@campus, omer@ee, ronen@ee\}.technion.ac.il}
}
\maketitle              % typeset the header of the contribution
\begin{abstract}
% \textcolor{red}{The abstract should briefly summarize the contents of the paper in 15--250 words.}
In a world abundant with diverse data arising from complex acquisition techniques, there is a growing need for new data analysis methods.
In this paper we focus on high-dimensional data that are organized into several hierarchical datasets. We assume that each dataset consists of complex samples, and every sample has a distinct irregular structure modeled by a graph.
The main novelty in this work lies in the combination of two complementing powerful data-analytic approaches: topological data analysis (TDA) and geometric manifold learning. 
Geometry primarily contains local information, while topology  inherently provides global descriptors. Based on this combination, we present a method for building an informative representation of hierarchical datasets. 
At the finer (sample) level, we devise a new metric between samples based on manifold learning that facilitates quantitative structural analysis.
At the coarser (dataset) level, we employ TDA to extract qualitative structural information from the datasets. 
We showcase the applicability and advantages of our method on simulated data and on a corpus of hyper-spectral images. We show that an ensemble of hyper-spectral images exhibits a hierarchical structure that fits well the considered setting. In addition, we show that our new method gives rise to superior classification results compared to state-of-the-art methods.
\end{abstract}
\section{Introduction}
\label{sec:introduction}

Modern datasets often describe complex processes and convey a mixture of a large number of natural and man-made systems. 
Extracting the essential information underlying such datasets poses a significant challenge, as these are often high-dimensional, multimodal, and without a definitive ground truth. Moreover, the analysis of such data is highly sensitive to measurement noise and other experimental factors, such as sensor calibration and deployment.
In order to cope with such an abundance, various data-analytic approaches have been developed,  aimed at capturing the structure of the data.
These approaches are often unsupervised, and are designed specifically to address the ``curse of dimensionality" in data.

In this paper we consider two complementing approaches for such ``structural" data analysis. Both of these approaches are based on the assumption that in high-dimensional real-world data, most of the information is concentrated around an intrinsic low-dimensional structure. Recovering the simplified underlying structure may reveal the true degrees of freedom of the data, reduce measurement noise, and facilitate efficient subsequent processing and analysis. 

The first approach we consider focuses on the \emph{geometry} of the data, and is called \emph{manifold learning} \cite{tenenbaum2000global,roweis2000nonlinear,belkin2003laplacian,Coifman2006}.  
The second approach focuses on the \emph{topology} of the data, and in known as \emph{topological data analysis} (TDA) \cite{carlsson_topology_2009,wasserman_topological_2016}. The key difference between these approaches is the distinction between local and precise phenomena (captured by geometry), and global qualitative phenomena (captured by topology).
Briefly, the goal in manifold learning is to obtain an accurate geometric representation of the manifold that best describes the data. This is commonly accomplished by approximating the Laplace-Beltrami operator of the manifold. On the other end, TDA  promotes the analysis of shapes and networks using qualitative topological features that are coordinate-free and robust under various types of deformations (e.g. the existence of holes). In a way, the topological descriptors are almost oblivious to the geometry and vice-verse. Our goal here is to take advantage of the strengths of each of these approaches, and combine them into a powerful geometric-topological framework.
%Recently, these methods have been applied successfully to various types of datasets \cite{adler_modeling_2017,nicolau_topology_2011,hiraoka_hierarchical_2016,chan_topology_2013,taylor_topological_2015,bendich_persistent_2016}.

Conceptually, the common thread between manifold learning and TDA is the premise that the true information underlying the data is encapsulated in the ``network" of associations within the data. Here lies another key difference between these two approaches.
Manifold learning methods traditionally represent such networks as graphs (i.e.~nodes and edges).
While graphs serve as a powerful model for various applications, this approach is limited since it can only capture \emph{pairwise} relationships between nodes.
However, it is highly conceivable that complex data and networks consist of much more intricate interactions, involving more than just two nodes at a time.
The methods developed in TDA focus on hypergraphs (simplicial complexes) that allow for high-order associations to be incorporated into the model \cite{dabaghian_topological_2012,giusti_clique_2015,lum_extracting_2013}. 

In this work, we propose to combine manifold learning and TDA in order to provide informative representations of high-dimensional data, under the assumption that they can be arranged into several \emph{hierarchical datasets} as follows. We assume that we have a collection of datasets, each consists of several complex samples, where each individual sample has a distinct irregular structure that can be captured by a weighted graph.
Such datasets arise in many applications from a broad range of fields such as cytometry and gene expression in bioinformatics \cite{giesen2014highly,edgar2002gene}, social and computer network analysis \cite{naitzat2020topology,leskovec2014snap}, medical imaging \cite{lamontagne2019oasis}, and geophysical tomography \cite{AVIRISweb}.
Following this hierarchy, our proposed method operates at two separate scales.

At the finer (sample) scale, we use an operator-theoretic approach to attach operators to  individual samples (graphs), pairs of samples, triplets, quadruplets, etc. These operators  quantitatively describe the structure of each sample separately, as well as the common structure across samples. Specifically, we  use the norm of these operators as a measure of similarity between samples, facilitating a transition from operator-theoretic analysis to affinity-based analysis.

%At the finer (sample) scale, we represent each we devise a new metric between samples based on manifold learning that facilitates quantitative structural analysis.
%This metric relies on an operator-theoretic approach, where we build an operator (matrix) to represent each sample (graph). Then, for each pair of samples, we build a composite operator based on the respective two operators, whose norm establishes the metric. 
%

At the coarser (dataset) scale, we employ TDA to extract qualitative information from the datasets. Concretely, we use \emph{persistent homology} \cite{edelsbrunner2008persistent,zomorodian2005computing} as a topological signature for each dataset. 
Persistent homology is a topological-algebraic tool that captures information about connectivity and holes at various scales. 
It is computed over the ensemble of samples contained in each dataset, which we model as a weighted simplicial complex where the weights are derived from the geometric operators computed at the finer scale.
The signature provided by persistent homology comes with a natural metric (the Wasserstein distance \cite{cohen-steiner_lipschitz_2010}), allowing us at the final stage to compare the structure of different datasets.

% OMER: I put the first sentence here below.
%The proposed methodology will be data-driven and ``model-free'' in the sense that it will not rely on prior knowledge or on rigid model assumptions. In the context of this project, this has paramount importance, since it circumvents the need to study each modality separately, as well as the ``hard wiring'' required for the fusion of different data sets.

%Topological Data Analysis (TDA) and Manifold learning gain increasing attention in recent works, which both facilitate compact informative representation and analysis for high-dimensional multi-modal data. However, both methods' shortcomings can be covered by the counterpart  method's advantage, by that to achieve exceptional data analytic approach.

%TODO: Lior - not sure if or where to write on other works.
%Another recent method that mesh both TDA and manifold learning techniques for dimension reduction is UMAP (Uniform Manifold Approximation and Projection) \cite{UMAP}.

To demonstrate the advantages of our method, we apply it to Hyper-Spectral Imaging (HSI) \cite{chang2003hyperspectral}.
HSI is a sensing technique aimed to obtain the electromagnetic spectrum at each pixel within an image, with the purpose of finding objects, identifying materials, or detecting processes.
We test our method on categorical hyper-spectral images \cite{AVIRISweb} and show that our unsupervised method accurately distinguishes between the different categories. 
In addition, we show that in a (supervised) classification task based on the attained (unsupervised) representation and metric, our method outperforms a competing method based on deep learning.

The main contributions of this work are:
(i) We introduce a powerful combination between geometry and topology, taking advantage of both local and global information contained in data.
(ii) We propose a method for analyzing hierarchical datasets, that is data-driven and ``model-free'' (i.e.~does not require prior knowledge or a rigid model).
(iii) We introduce a new notion of affinity between manifolds, quantifying their commonality.

%The paper is organized as follows. In section \ref{sec:model} we formulate the problem. 
%In section \ref{sec:background}, we present preliminaries on manifold learning and TDA. Section \ref{sec:method} provides a detailed description of the proposed method. In section \ref{sec:simulation}, we present a simulation study, and in section \ref{sec:application}, we show application to HSI. 
%Lastly, in section \ref{sec:conclusion} we conclude and discuss directions for future work.

% More concretely, 

%  we use manifold learning techniques to facilitate quantitative structural analysis for each individual sample (modeled as a graph)

\section{Problem Formulation}
\label{sec:model}

In this work, we consider the following hierarchical structure. At the top level, we have a collection of $N_D$ \emph{datasets}
\[{\cal D} = \{D_1,\ldots,D_{N_D}\}.\]
These datasets may vary in size, shape and origin. However, we assume that they all have the same prototypical structure, as follows. Each dataset $D$
consists of a collection of $N$ \emph{samples}
\[
D = \{S_{1}, \ldots, S_{N}\}, 
\]
and each sample is a collection of $L$ \emph{observations}
\[
S_{i} = \{x_{i,1} \ldots, x_{i,L}\}, \ 1 \le i \le N.
\]
Note that $N$ and $L$ may vary between datasets (for simplicity we omit the dataset index), but all the samples within a single dataset are of the same size $L$.

Next, we describe the structure of a single dataset $D$. 
Let $\{\M_\ell\}_{\ell=1}^M$ be a set of latent manifolds, and let $\Pi$ be their product
\begin{equation}
\label{eqn:manifold_product}
\Pi=\M_{1} \times \cdots \times  \M_{M}.
\end{equation}
We use $\Pi$ as a model for the common hidden space underlying the dataset $D$. Let ${\cal X} = \{x_1,\ldots, x_L\}$ be a set of points sampled from $\Pi$, where each point can be written as a tuple $x_j = (x_j^{(1)},\ldots, x_j^{(M)})$ and $x_j^{(\ell)}\in \mathcal{M}_{\ell}$ for $\ell=1,\ldots,M$.

Our main assumption here is that all samples $S_i$ are generated by the same set $\cal X$, while each sample contains information only about a subset of the manifolds in the product \eqref{eqn:manifold_product}.
The subset of manifolds corresponding to a sample $S_i$ is represented by a tuple of indices $I_i = (\ell_1^{(i)},\ldots,\ell_m^{(i)})$ ($m \le M$), and a product manifold
\begin{equation}\label{eqn:subset_product}
\Pi_{I_i} = \M_{\ell_1^{(i)}} \times \cdots \times  \M_{\ell_m^{(i)}}.
\end{equation}
For convenience, for each $x\in \Pi$, we define the projection
\[
x_{I_i} = (x^{(\ell_1^{(i)})},\ldots,x^{(\ell_m^{(i)})}) \in \Pi_{I_i}.
\]
Next, for each sample $S_i$ (and a corresponding subset $I_i$) we assume there is a function $g_i : \Pi_{I_i} \to {\cal O}_i$, for some target metric space ${\cal O}_i$. We define the observation function $f_i:\Pi\to {\cal O}_i$ as 
\[
    f_i(x) = g_i(x_{I_i}) + \xi_i,
\]
where $\xi_i \in {\cal O}_i$ denotes a random independent observation noise.
Finally, the sample $S_i$ is defined as
\[
    S_i = \{x_{i,1},\ldots,x_{i,L}\} =  \{f_i(x_1),\ldots,f_i(x_L)\}.
\]
In other words, each sample $S_i$ in the dataset $D$ reveals partial and noisy information about $\Pi$.
As stated earlier, we assume that each of the datasets $D_k\in {\cal D}$ is generated by the model described above. However, the manifold $\Pi$, the functions $f_i,g_i$, and the parameters $L,M,N,I_i$ and $\xi_i$ may differ between datasets.
Note that within a single dataset there is a correspondence between all samples $S_{1},\ldots, S_{N}$, as they are generated by the same set of realizations ${\cal X}$.

\begin{figure}[t]
    \centering
    \includegraphics[width=0.7\textwidth]{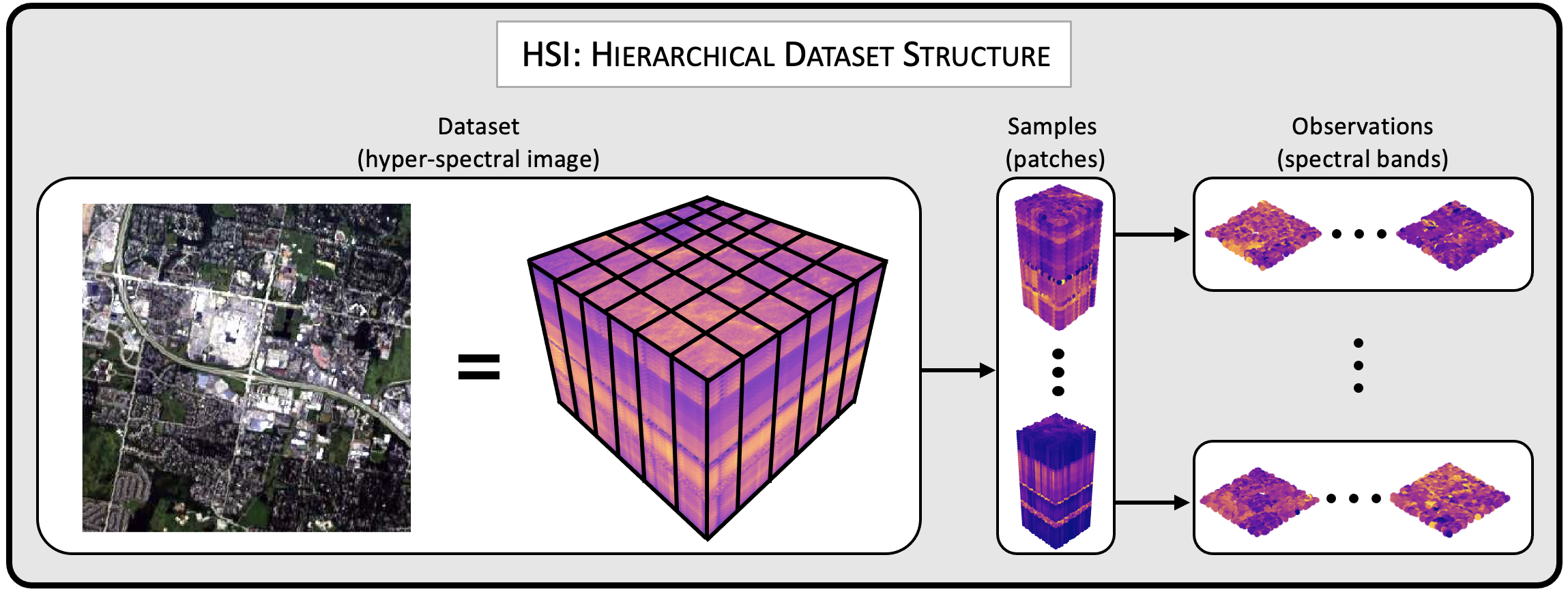}
    \caption{The specification of the considered hierarchical structure for HSI.}
    \label{fig:HSI_hie}
\end{figure}

In the context of HSI, the hierarchical structure described above is as follows.
Each hyper-spectral image is viewed as a single dataset, the full spectrum of a single patch as a sample, and different spectral bands within a patch as observations (see Figure \ref{fig:HSI_hie} and Appendix \ref{sec:appendix_diffusion_maps}).

\section{Background}
\label{sec:background}
In this section, we review some preliminaries required to describe our proposed method. Section  \ref{sec:Diffusion_Opertators} presents the diffusion and the alternating diffusion operators.
Section \ref{sec:TDA} provides a brief introduction to persistent homology. 

\subsection{Multiple manifold learning and diffusion operators}
\label{sec:Diffusion_Opertators}

Manifold learning is a class of unsupervised nonlinear data-driven methods for discovering the geometric structure underlying high dimensional data \cite{tenenbaum2000global,roweis2000nonlinear,belkin2003laplacian,Coifman2006}. The main assumption in manifold learning is that high-dimensional data lie on a hidden lower-dimensional manifold.

One of the notable approaches in manifold learning is \emph{diffusion maps} \cite{Coifman2006}, in which diffusion operators built from data are shown to approximate the Laplace-Beltrami operator.
This differential operator contains all the geometric information on the manifold \cite{berard1994embedding}, and thus its approximation provides means to incorporate geometric concepts such as metrics and embedding into data analysis tasks \cite{jones2008manifold}.
%diffusion processes, which in turn designed to capture the structure of the sources of variability in data. The diffusion operator attempt to capture the probability density of data points on the underlying manifold, only through the observation space. As shown in \cite{belkin2003laplacian}, by using Gaussian kernel $k(x, x') = \exp ( - \frac{\norm{x - x'}^2}{\epsilon} )$ where $x,x'$ are observations, if the density of points is uniform, then as $\epsilon \rightarrow 0$, one is able to approximate the Laplace-Beltrami operator on the latent manifold. Following that, In \cite{coifman2005geometric} they approximate diffusion process of data on manifolds. The concept key is that the diffusion operator is a ``local'' kernel which is a Markov kernel that hold the information on data distribution on the latent manifold.
%
In \cite{lederman2018learning,talmon2017latent}, an extension of diffusion maps for multiple datasets termed `alternating diffusion' was introduced. This extension, based on the product of diffusion operators, was shown to recover the manifold structure common to multiple datasets.
In this work, we utilize a variant of alternating diffusion, proposed in \cite{shnitzer2019recovering}, which is briefly described in the remainder of this subsection.

Consider two diffeomorphic compact Riemannian manifolds without a boundary, denoted by $(\M_{1},g_1)$ and $(\M_{2},g_2)$, and a diffeomorphism $\phi: \M_{1} \rightarrow \M_{2}$.
For each manifold ${\cal M}_\ell$ ($\ell=\{1,2\}$) and a pair of samples $x,x'\in \mathcal{M}_{\ell}$, let $k_\ell(x, x')$ be a Gaussian kernel based on the distance induced by the metric $g_\ell$ with a kernel scale $\epsilon_\ell > 0$.
Define $d_\ell(x)=\int_{\M_\ell} k_\ell(x,x') \mu_\ell(x') dv_\ell(x')$, where $v_\ell(x)$ is the volume measure, and $\mu_\ell(x')$ is the density function of the samples on $\M_\ell$.
Consider two kernel normalizations: $a_\ell(x,x') = \frac{k_\ell(x,x')}{d_\ell(x)}$ and $b_\ell(x,x') = \frac{k_\ell(x,x')}{d_\ell(x')}$.
Based on the normalized kernel $a_\ell(x,x')$, define the \emph{forward diffusion operator} by $A_\ell f(x) = \int_{\M_\ell} a_\ell(x,x') f(x') \mu_\ell(x') dv_\ell(x')$ for any function $f\in C^{\infty}(\M_\ell)$. Similarly, based on $b_\ell(x,x')$, define the \emph{backward diffusion operator} by $B_\ell f(x) = \int_{\M_\ell} b_\ell(x,x') f(x') \mu_\ell(x') dv_\ell(x')$. 
In a single manifold setting, it is shown that when $\epsilon_\ell \rightarrow 0$, $A_\ell$ converges to a differential operator of an isotropic diffusion process on a space with non-uniform density $\mu_\ell(x)$, and $B_\ell$ converges to the backward Fokker-Planck operator, which coincides with the Laplace-Beltrami operator when the density $\mu_\ell(x)$ is uniform \cite{nadler2006diffusion}.

Next, for two manifolds, consider the following $C^\infty(\M_1) \rightarrow C^\infty(\M_1)$ composite operators,
\[
    G f(x) = \phi^* A_{2} (\phi^*)^{-1} B_{1} f(x),
\]
and
\[
    H f(x) = A_{1} \phi^* B_{2} (\phi^*)^{-1} f(x)
\]
for any function $f\in C^\infty(\M_{1})$, where $\phi^*: C^\infty(\M_{2}) \rightarrow C^\infty(\M_{1})$ denotes the pullback operator from $\M_{2}$ to $\M_{1}$ and $(\phi^*)^{-1}$ denotes the push-forward from $\M_{2}$ to $\M_{1}$, both corresponding to the diffeomorphism $\phi$.

In \cite{lederman2018learning,talmon2017latent}, it was shown that the two composite operators $G$ and $H$ recover the common structure between $\M_1$ and $\M_2$ and attenuate non-common structures that are often associated with noise and interference.
In \cite{shnitzer2019recovering}, the following symmetric alternating diffusion operator was introduced
\[
    {S}_{1, 2} f(x)= \frac{1}{2} \Big( G f(x) + H f(x) \Big),
\]
which, in addition to revealing the common structure as $G$ and $H$, has a real spectrum -- a convenient property that allows to define spectral embeddings and spectral distances. 

%Note, for sufficiently small kernel scales $\epsilon$ and $\mu$ is smooth enough, the asymptotic expansion of the operator $\bm{S}_{1,2}$ include a summation of two Laplace-Beltrami operators, corresponding to the two considered manifolds, $\M_1$ and $\M_2$. As shown in \cite{Froyland_2015,Froyland2020}, the summation of two Laplace-Beltrami operators, from two different ``time''-instances, under the assumption of a uniform density is relates to the dynamic Laplacian of some dynamical system. Moreover, the dynamic Laplacian reveals coherent sets in dynamical systems, representing common system behavior in different time-instances, by that the operator $\bm{S}_{1,2}$ represents the commonality between the manifolds.

In practice, the operators defined above are approximated by matrices constructed from  finite sets of data samples.
Let $\{(x_i^{(1)},x_i^{(2)})\}_{i=1}^L$ be a set of $L$ pairs of samples from $\M_{1}\times\M_{2}$, such that $x^{(2)}_i = \phi(x^{(1)}_i)$.
% Assume that the samples are embedded in two different high-dimensional observation spaces.
For $\ell=\{1,2\}$, let $\bm{W}_\ell$ be an $L \times L$ matrix, whose $(i,i')$-th element is given by $W_{\ell}(i,i') = k_{\ell}(x_i^{(\ell)}, x_{i'}^{(\ell)})$. Let $\bm{Q}_\ell = \text{diag}(\bm{W}_{\ell}\bm{1})$ be a diagonal matrix, where $\bm{1}$ is a column vector of all ones.
The discrete counterparts of the operators $A_\ell$ and $B_\ell$ are then given by the matrices $\bm{K}_{\ell} = (\bm{Q}_{\ell} )^{-1} \bm{W}_{\ell}$ and $\bm{K}^T_{\ell}$, respectively, where $(\cdot)^T$ is the  transpose operator.
Consequently, the discrete counterparts of the operators $G$ and $H$ are
$\bm{G} = \bm{K}_{2} \bm{K}_{1}^T$ and $\bm{H} = \bm{K}_{1} \bm{K}_{2}^T$, respectively.
The symmetric matrix corresponding to the operator $S_{1,2}$ is then
\[
    \bm{S}_{1,2} = \bm{G} + \bm{H}.
\]
% TODO: moved from intro, should be merged here:
% Unsupervised Manifold Learning is a class of nonlinear data-driven methods for discovering the underlying manifolds in high dimensional data \cite{diffusion_maps,ISOMAP,LLE,Laplacian_Eig}. One common assumption behind manifold learning is that data observations of some phenomena lie on a hidden lower-dimensinal manifold. This assumption give solid ground for the incorporation of geometric concepts such as metrics and embedding that facilitate to data analysis tasks.
% One way to extract the latent manifold from data samples is by construction of graph. Typically, the data samples form the graph vertices and the edges are determined according to some similarity notion that is usually application oriented. To name few domains that manifold learning is well performing is dynamical systems \cite{dynamical_systems_1,dynamical_systems_2}, multi-modal data \cite{multi_modal_1} and multi-view data \cite{multi_view}. The major shortcoming of manifold learning is that it is rather quantitative than qualitative, by learning the fine interaction between data samples can mislead in some cases to get a general analysis of datasets.

\subsection{Simplicial complexes and persistent homology}
\label{sec:TDA}

At the heart of the topological layer in our proposed method, we will use (abstract) simplicial complexes to represent the structure of a dataset. 
Briefly, a simplicial complex is a discrete structure that contains vertices, edges, triangles and higher dimensional simplexes (i.e.~it is a type of hypergraph). This collection has to be closed under inclusion -- for every simplex we must also include all the faces on its boundary. See Figure \ref{fig:ghrist} for an example.

One of the many uses of simplicial complexes is in network modeling.
While graphs  take into account pairwise interactions between the nodes, simplicial complexes allow us to include information about the joint interaction of triplets, quadruplets, etc. We will use this property later, when studying the structure of samples within a dataset.

{\bf Homology}
is an algebraic structure that describes the shape of a topological space. Loosely speaking, for every topological space (e.g.~a simplicial complex) we  can define a sequence of vector spaces $H_0,H_1,H_2,\ldots$ where $H_0$ provides information about connected components, $H_1$ about closed loops surrounding holes, $H_2$ about closed surfaces enclosing cavities. Generally, we say that $H_k$ provides information about \emph{$k$-dimensional cycles}, which can be thought of as  $k$-dimensional surfaces that are ``empty" from within. We describe homology in more detail in Appendix \ref{sec:appendix_persistent_homology}.
In this work we will mainly use $H_0$ and $H_1$, i.e.~information about connectivity and holes. However, the framework we develop can be used with any dimension of homology.

\begin{figure}[t]
    \centering
    \includegraphics[width=0.7\textwidth]{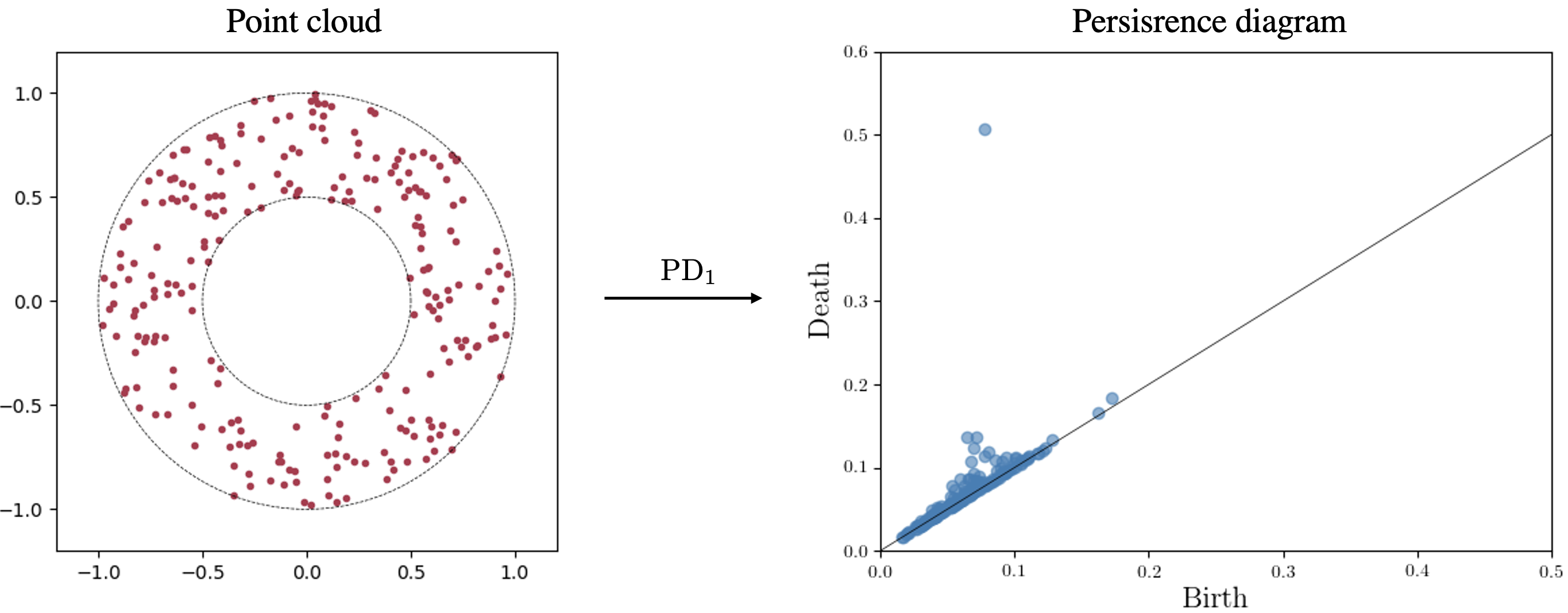}
    \caption{Persistence diagram. Left: a point cloud generated in an annulus with a single hole ($1$-cycle). The filtration used is the union of balls around the points, for an increasing radius. Right:  persistence diagram for $1$-cycles. The birth/death axes represent radius values. The single feature away from the diagonal represents the hole of the annulus, while  other cycles are considered  ``noise".}
    \label{fig:pers}
\end{figure}

{\bf Persistent homology}
is one of the most heavily used tools in TDA \cite{edelsbrunner2008persistent,zomorodian2005computing}. It can be viewed as a \emph{multi-scale} version of homology, where instead of considering the structure of a single space, we track the evolution of cycles for a nested sequence of spaces, known as a \emph{filtration}. As the spaces in a filtration grow, cycles of various dimensions may form (born) and  later get filled in (die). The $k$-th persistent homology, denoted $\mathrm{PH}_k$, keeps a record of the birth-death process of $k$-cycles. Commonly, the information contained in $\mathrm{PH}_k$ is summarized using a \emph{persistence diagram} -- a set of points in $\R^2$ representing all the (birth,death) pairs for cycles in dimension $k$, and denoted $\mathrm{PD}_k$ (see Figure \ref{fig:pers}). The motivation for using persistent homology is that it allows us to consider cycles at various scales, and identify those that seem to be prominent features of the data.

In order to compare between persistence diagrams,  we will employ the Wasserstein distance \cite{cohen-steiner_lipschitz_2010}
defined as follows. Suppose that $\mathrm{PD}$ and $\mathrm{PD}'$ are two persistence diagrams, then the $p$-Wasserstein distance is defined as
\begin{equation}
    d_{W_p}(\mathrm{PD},\mathrm{PD}') = \inf_{\phi:\widetilde{\mathrm{PD}} \rightarrow \widetilde{\mathrm{PD}}'} \Big( \sum_{\alpha \in \widetilde{\mathrm{PD}}} \norm{\alpha - \phi(\alpha)}^p\Big)^{\frac{1}{p}},
\end{equation}
where $\widetilde{\mathrm{PD}}$ is an augmented version of $\mathrm{PD}$ that also includes the diagonal line $x=y$ (and the same goes for $\widetilde{\mathrm{PD}}'$). This  augmentation is taken in order to allow cases where the $|\mathrm{PD}|\ne |\mathrm{PD}'|$. In other words, the Wasserstein distance is based on an optimal matching between features in $\mathrm{PD}$ and $\mathrm{PD}'$.

\section{Proposed Method}
\label{sec:method}
\begin{figure}[t]
    \centering
    \includegraphics[width=0.7\textwidth]{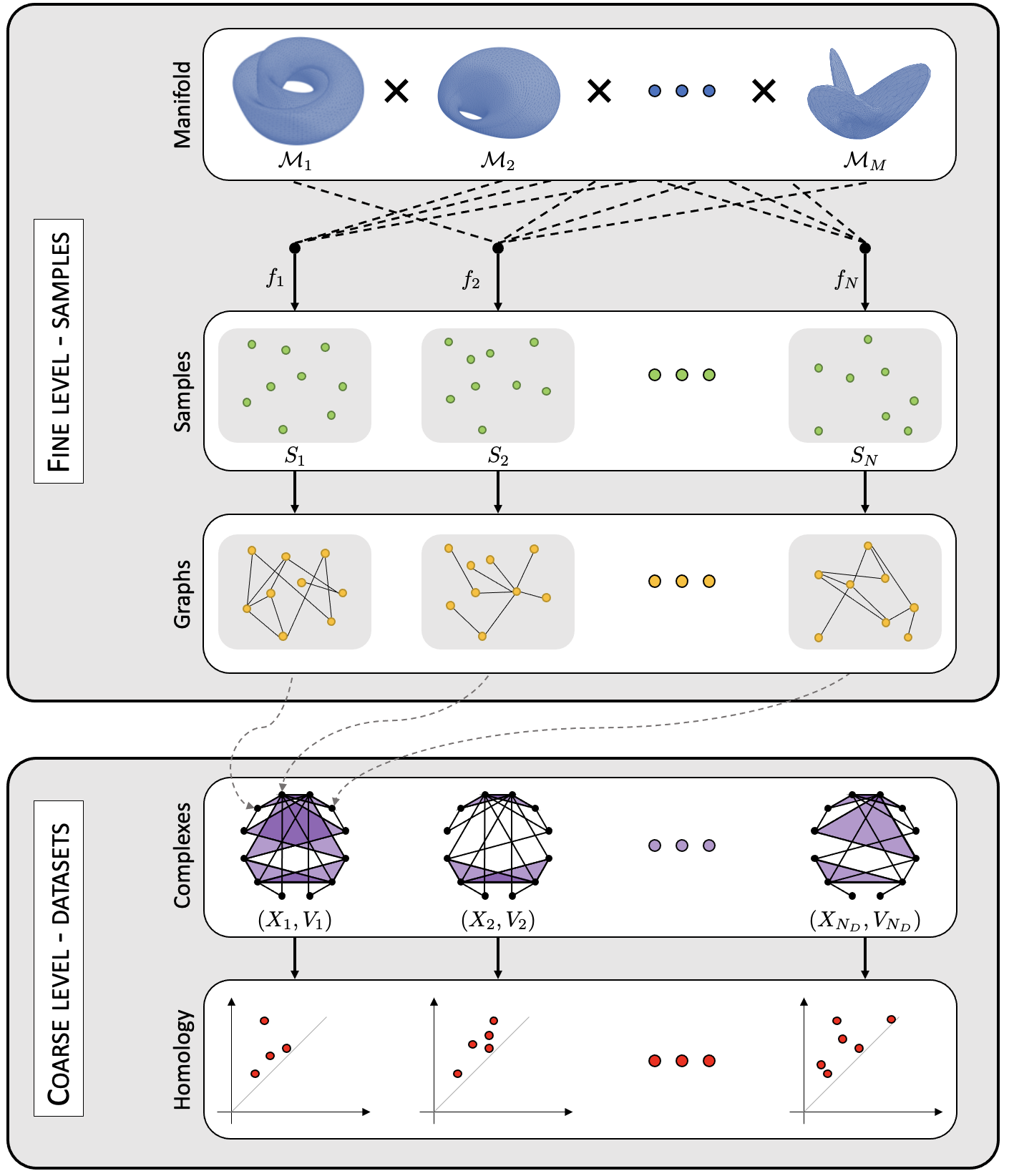}
    \caption{Method outline. Each dataset (bottom part) is represented as  weighted simplicial complex, where the weights are calculated using the alternating diffusion operator between the sample graphs (see \eqref{eqn:alternating}, \eqref{eq:threeway_ad}).
     The output is a set of persistence diagram in the last row, where each diagram summarizes a single dataset. We can compare the datasets using the Wasserstein distance \eqref{eqn:d_dataset}.}
    \label{fig:outline}
\end{figure}

Recall the hierarchical dataset structure presented in Section \ref{sec:model}.
The processing method we propose for such datasets is hierarchical as well. At the fine level, each sample $S_{i}$ is treated as a weighted graph, which we analyze geometrically using a diffusion operator. At the coarse level, each dataset $D$ is considered as a weighted simplicial complex, from which we extract its persistent homology, enabling us to compare between different datasets using the Wasserstein distance. Figure \ref{fig:outline} summarizes this pipline.

The motivation for this analysis is the following. At the fine level, we use geometry in order to capture the detailed structure of a sample. Since all the samples within a dataset $D$ are assumed to be generated by the same set of realizations ${\cal X}\subset \Pi$, their geometry provides a solid measure of inter-sample similarity.
Conversely, at the coarse level, the geometry of different datasets can be vastly different. Thus, in order to compare datasets, we propose to use topology as an informative representation of the global qualitative structure, rather than geometry.

\subsection{The sample diffusion operator}
\label{subsec:sample_diffusion_operator}
We treat each sample $S_{i}$ as a weighted graph, with weights calculated using a Gaussian kernel, forming an affinity matrix $\bm{W}_i \in \mathbb{R}^{L \times L}$, whose $(j_1,j_2)$-th element is given by
\begin{equation}\label{eq:kernel}
    W_{i}(j_1,j_2) = \exp\left(-\frac{d^2_{i}(x_{i,j_1}, x_{i,j_2})}{\epsilon}\right),
\end{equation}
where $d_{i}$ is a distance suitable for the observation space ${\cal O}_{i}$.
% TODO: index i here and \ell in Sec 3.1.

Next, following \cite{Coifman2006}, we apply a two-step normalization. The first step is designed to handle a possibly non-uniform density of data points on the manifold. Let $\bm{Q}_i=\text{diag}(\bm{W}_i\bm{1})$ be a diagonal matrix that approximates the local densities of the nodes, where $\bm{1}$ is a column vector of all ones, and define
\begin{equation}\label{eq:first_normalization}
    \widetilde{\bm{W}}_i = \bm{Q}_i^{-1} \bm{W}_i \bm{Q}_i^{-1}.
\end{equation}
In the second step, we build anther diagonal matrix $\widetilde{\bm{Q}}_i=\text{diag}(\widetilde{\bm{W}}_i\bm{1})$, and form the following stochastic matrix
\begin{equation}\label{eq:diffusion_operator}
    \bm{K}_i = \widetilde{\bm{Q}}_i^{-1}\widetilde{\bm{W}}_i,
\end{equation}
which is called the \emph{diffusion operator} of  sample $S_i$.
We note that the construction of $\bm{K}_i$ is similar to the construction described in Section \ref{sec:Diffusion_Opertators} that follows \cite{shnitzer2019recovering} with only one difference -- the first normalization that copes with non-uniform sampling.

%%%%%%%%%%%%%%%%%%%%%%%%%%%%%%%%%%%%%%%%%%%%%%%%%%%%%%%%%%%%%%%%%%%%%%%%%%%%
\subsection{The dataset simplicial complex}
\label{subsec:dataset_sim_comp}
We construct a weighted simplicial complex for every dataset $D$, whose vertex set correspond to the samples $\{S_{i}\}_{i=1}^{N}$. We assume that the simplicial complex is given (and depends on the problem at hand but not on the data), and we only need to determine the weights on the simplexes. 

Considering our model in \eqref{eqn:manifold_product} and \eqref{eqn:subset_product}, we propose weights that are inversely correlated with the number of common hidden variables between the samples. Denote by $V$ the weight function for the simplexes representing dataset $D$. Ideally, for any $d$-dimensional simplex $\sigma = [i_1,\ldots,i_{d+1}]$ we want to have
\begin{equation}\label{eqn:heuristic}
    V(\sigma) = U(|I_{i_1}\cap\cdots\cap I_{i_{d+1}}|),
\end{equation}
where $I_{i}$ is the tuple of indexes corresponding to sample $S_i$ (see Section \ref{sec:model}), 
and where $U$ is a decreasing function. While this condition cannot hold in a strict sense (mainly due to observation noise), the method we devise below provides a close approximation.

We start with the edges. Let $\bm{K}_{i_1},\bm{K}_{i_2}$ be a pair of diffusion operators for samples $S_{i_1},S_{i_2}\in D$ (see Section \ref{subsec:sample_diffusion_operator}). As described in Section \ref{sec:Diffusion_Opertators}, the work in \cite{shnitzer2019recovering}, based on the notion of alternating diffusion \cite{lederman2018learning}, showed that one can reveal the common manifold structure between $S_{i_1}$ and $S_{i_2}$ by considering the symmetric alternating diffusion operator
\begin{equation}\label{eqn:alternating}
    \bm{S}_{i_1,i_2} = \bm{K}_{i_1} \bm{K}_{i_2}^T + \bm{K}_{i_2} \bm{K}_{i_1}^T.
\end{equation}
As a heuristic, we propose to set the weight function $V$ to be the inverse of the Frobenius norm, i.e.
\begin{equation}\label{eqn:weight_edge}
    V([i_1,i_2]) = \norm{\bm{S}_{i_1,i_2}}_F^{-1}.
\end{equation}
The rationale behind this heuristic stems from the common practice in kernel methods. Typically, the eigenvalues of the kernel are used for evaluating the dominance of the component represented by the corresponding eigenvectors. 
Indeed, using the spectral distance was proposed in \cite{rajendran2016data} in a setting where the samples form individual graphs, as in the current work.
Here, we follow the same practice but with a kernel that captures only the common components. 
We will show empirically in Section \ref{sec:simulation} that indeed $V([i_1,i_2])$ inversely correlates with $|I_{i_1}\cap I_{i_2}|$, as desired.

Next, we consider triangles in our complex. In a similar spirit to \eqref{eqn:alternating}, we define the three-way symmetric alternating diffusion operator by
\begin{equation}\label{eq:threeway_ad}
    \bm{S}_{i_1,i_2,i_3} = \bm{S}_{i_1,i_2} \bm{K}_{i_3}^T + \bm{K}_{i_3}\bm{S}_{i_1,i_2}
    +\bm{S}_{i_2,i_3} \bm{K}_{i_1}^T + \bm{K}_{i_1} \bm{S}_{i_2,i_3} 
    +\bm{S}_{i_1,i_3} \bm{K}_{i_2}^T + \bm{K}_{i_2}\bm{S}_{i_1,i_3}.
\end{equation}
The weight function of the corresponding triangle is then set as
\begin{equation}\label{eqn:weight_triangle}
    V([i_1,i_2,i_3]) = \norm{\bm{S}_{i_1,i_2,i_3}}_F^{-1}.
\end{equation}
In Section \ref{sec:simulation} we also show empirically that $V([i_1,i_2,i_3])$ inversely correlates with $|I_{i_1}\cap I_{i_2}\cap I_{i_3}|$. In particular, we have $V([i_1,i_2]) \le V([i_1,i_2,i_3])$ for all $i_1,i_2,i_3$, which is required in order to have a filtered complex.
% TODO [ronen]: is it immediate? can we show it in an appendix?

In a similar spirit, one can define $V$ for simplexes of any dimension. However, for the simulation and application we consider here, edges and triangles suffice.
%
% TODO [ronen]: i want to include the following text here:
% It is worthwhile noting that \eqref{eqn:weight_edge} and \eqref{eqn:weight_triangle} allow us to transition from operator-theoretic analysis to affinity-based analysis, which is described next.

\subsection{Topological distance between datasets}

The proposed pipeline concludes with a numerical measure of structural similarity between two datasets $D$ and $D'$.
Recall that the output of the previous section are weighted simplicial complexes, denoted by the pairs $(X,V)$ and $(X',V')$, where $X$ and $X'$ are complexes and $V$ and $V'$ are the weight functions. We use each weight function to generate  a filtration that in turn serves as the input to the persistent homology computation (see Section \ref{sec:TDA}). The filtration we take is the \emph{sublevel} set filtration $\{X_v\}_{v\in \R}$, where $X_v = \{\sigma : V(\sigma) \le v\}$. Considering the weights constructed in \eqref{eqn:weight_edge} and \eqref{eqn:weight_triangle} , this implies that simplexes that represent groups of samples that share more structure in common will appear earlier in the filtration.

Let $\mathrm{PD}_{k}$ and $\mathrm{PD}_k'$ be the $k$-th persistence diagrams  of $(X,V)$ and $(X',V')$, respectively. We can then compare the topology of two datasets by calculating the Wasserstein distance
\begin{equation}\label{eqn:d_dataset}
d_{\mathrm{dataset}}(D, D') = d_{W_p}(\mathrm{PD}_{k},\mathrm{PD}_{k}').
\end{equation}
The choice of $p$ and $k$ depends on the application at hand.
The entire pipeline is summarized in Algorithm \ref{algo:TDA}. Note that it is currently described for $k=\{0,1\}$, but once the weight function $V$ in Subsection \ref{subsec:dataset_sim_comp} is extended beyond edges and triangles to higher orders, the algorithm can be extended for $k\ge2$ as well. 

\begin{algorithm}[t]
    \SetAlgoLined
    \SetKwInOut{Input}{Input}
    \SetKwInOut{Output}{Output}
    \SetKwInOut{Parameter}{Parameters}
    \Input{Two hierarchical datasets: $D$ and $D'$}
    \Output{Distance between the datasets: $d_{dataset}(D,D')$}
    \Parameter{$k=\{0,1\}$ (homology degree), $p$ (Wasserstein distance order), $\epsilon$ (kernel scale)}
    \begin{enumerate}
        \item Construct a simplicial complex $X$ for each dataset as follows:
            \begin{enumerate}
                \item For each sample $S_i$, $i=1, ... N$ compute the diffusion operator
                    % \begin{enumerate}
                    %     %\item Construct the affinity matrix $\bm{W}_i$ with scale parameter $\epsilon$ according to \eqref{eq:kernel}
                    %     %\item Compute the normalized affinity matrix $\widetilde{\bm{W}}_i$ according to  \eqref{eq:first_normalization}
                    %     \item Compute the diffusion operator
                    $\bm{K}_i$, with scale parameter $\epsilon$ according to \eqref{eq:kernel}--\eqref{eq:diffusion_operator}
                    % \end{enumerate}
                \item For all edges $\forall (i_1,i_2) \in 1, \ldots L$:
                \begin{enumerate}
                    \item Compute the symmetric alternating diffusion operator $\bm{S}_{i_1,i_2}$ according to \eqref{eqn:alternating}
                    \item Set the weight function: $V([i_1, i_2]) = \norm{\bm{S}_{i_1,i_2}}_F^{-1}$
                \end{enumerate}
                \item For all triangles $\forall (i_1,i_2,i_3) \in 1, \ldots L$:
                \begin{enumerate}
                    \item Compute the three-way symmetric alternating diffusion operator $\bm{S}_{i_1,i_2,i_3}$ according to \eqref{eq:threeway_ad}
                    \item Set the weights of $V([i_1, i_2, i_3]) = \norm{S_{i_1,i_2,i_3}}_F^{-1}$
                \end{enumerate}
            \end{enumerate}
        \item Compute the $k$-th persistence diagram for the weighted complexes $(X,V)$ and $(X',V')$, corresponding to $D$ and $D'$, respectively.
        \item Compute the distance between the persistence diagrams: $d_{W_p}(\mathrm{PD}_k,\mathrm{PD}_k')$
    \end{enumerate}
    \caption{A geometric-topological distance between two datasets}
    \label{algo:TDA}
\end{algorithm}

\section{Simulation Study}
\label{sec:simulation}
In this section, we test the proposed framework on a toy problem, where we can manipulate and examine all the ingredients of our model and method. 

We start with the description of a single dataset $D$.
Revisiting the notation in Section \ref{sec:model}, we assume that the latent manifold for each dataset is of the form $\Pi = {\cal M}_1\times \cdots\times{\cal M}_M$, where each $\mathcal{M}_{\ell}$ is a circle of the form
\begin{equation}
    \M_\ell = \{ (\cos(\theta_\ell), \sin(\theta_\ell)) \, \lvert \, 0 \leq \theta_\ell < 2\pi \} .
\end{equation}
In other words, $\Pi$ is an $M$-dimensional torus. In this case, the latent realization set ${\cal X} = \{x_{1},\ldots,x_{L}\}$ is a subset of $\R^{2M}$. 
We generate ${\cal X}$ by taking a sample of iid variables $\theta_{\ell}\sim U[0,2\pi)$ for  $\ell=1, \ldots, M$.
For all the samples $S_{i}$ ($1\le i \le N$) we take $|I_{i}| = 3$, and the observation space is then ${\cal O}_{i} =\R^6$.
For every $x\in \Pi$, we define
\[
    g_{i}(x_{I_i}) = \\
    \begin{pmatrix}
                R_1\cos(\theta_{\ell_1}),
                R_1\sin(\theta_{\ell_1}), 
                R_2\cos(\theta_{\ell_2}), 
                R_2\sin(\theta_{\ell_2}),
                R_3\cos(\theta_{\ell_3}), 
                R_3\sin(\theta_{\ell_3})
    \end{pmatrix}
\]
where $I_{i} = (\ell_1,\ell_2,\ell_3)$ are the indexes of the subset of manifolds viewed by sample $S_{i}$. The radii $R_1,R_2,R_3$ are generated uniformly at random in the interval $[1,R_{\max}]$, for each $i$ independently. The indexes $\ell_1,\ell_2,\ell_3$ are also chosen at random.
Finally, the sample observation function is given by $f_{i}(x) = g_{i}(x_{I_i}) + \xi_i$, where $\xi_i \sim \mathrm(0, \sigma_i^2 I)$ (independent between observations).

For the generation of all datasets, we use $N = 40$, and $L = 200$. The torus dimension $M$ varies between $3$ and $30$ across the datasets. 
As $M$ increases, the chances that the pair of samples $(S_i,S_j)$ has underlying circles in common decreases. Subsequently, the connectivity of the simplicial complex of the respective dataset decreases. Thus, $M$ strongly affects the affinity between the datasets.
% [omer] Is "affinity" ok here?

\begin{figure}[t]
    \centering
    \includegraphics[width=\textwidth]{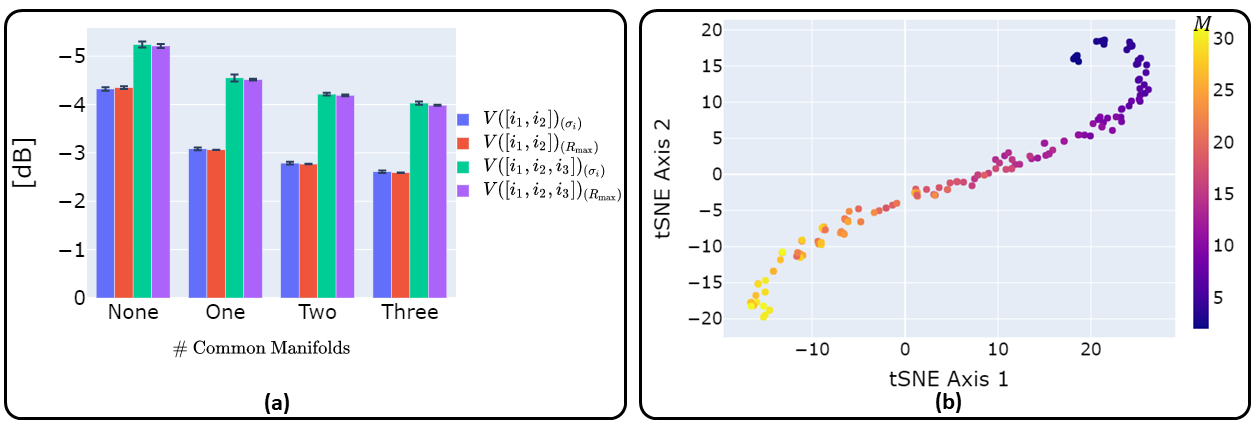}
    \caption{(a) The mean and standard deviation of the  weight function $V$ (for edges and triangles). We take $20$ random realizations, as well as varying $R_{\text{max}}$ between $1-15$ and $\sigma_i$ between $0.001-1000$. Note that we plot the value $\text{log}(1-V)$, and the $y$ axis is flipped, so this is indeed a monotone-decreasing behavior.
    (b) t-SNE embedding based on the Wasserstein distance for $H_1$ (holes), where the color indicates the size of the pool of underlying manifolds $M$.}
    \label{fig:sim}
\end{figure}

%\subsection{Testing the weight heuristic}
In Section \ref{sec:method} we argued that the weight function $V$ defined in \eqref{eqn:weight_edge} and \eqref{eqn:weight_triangle} is roughly decreasing in the number of common variables \eqref{eqn:heuristic}.
Here, we provide an experimental evidence for that heuristic. 
In Figure \ref{fig:sim}(a), we plot the mean and standard deviation of the weight function (in dB) as a function of the number of common indices (manifolds/circles). 
We calculate $V$ across $20$ realizations, 
% [omer] How many realizations here?
and across various choices of $R_{\max}$ ($1-15$) and $\sigma_i$ ($0.001-1000$). The results clearly indicate a monotone decreasing relationship between $V$ and the number of common manifolds. In addition, robustness to noise ($\sigma_i$) and to the particular observation space ($R_{\text{max}}$) is demonstrated.

%\subsection{Distance between datasets}
% \begin{figure}
%     \centering
%     \includegraphics[width=0.5\textwidth]{Figs/Sim_tSNE.eps}
%     \caption{t-SNE embedding based on the Wasserstein distance for $H_1$ (holes), the color indicate the underlying manifolds pool size in the dataset.}
%     \label{fig:sim_tsne}
% \end{figure}

% [[[When $k=3$, in all samples, the underlying circles must be identical because $I$ picks $3$ circles from a pool of size $3$. 
% While $k$ is increasing, the chance that two subsets $I,J \in \mathcal{I}_k$ have underlying circles in common is decreasing.
% Therefore, $k$ control the general structure in the datasets $\bm{\mathcal{P}}_{k}$ and determined the distance between datasets:
% \begin{equation}
%     d_{dataset}\big( \bm{\mathcal{P}}_{k_1}, \bm{\mathcal{P}}_{k_2} \big) \propto \abs{k_1-k_2}
% \end{equation}
% \footnote{I didn't understand at all the point of this paragraph, plus I don't think (11) is correct. I'll leave it for now, but I would just remove it}.]]]

For each $3\le M \le 30$ we generate $5$ datasets with $R_{\max}=15$ and $\sigma_i = 0.1$, so that overall there are $N_D = 140$ datasets. For each dataset, we follow Algorithm \ref{algo:TDA} and calculate the topological distance $d_{\mathrm{dataset}}$ \eqref{eqn:d_dataset} using $k=1$ (holes) and $p=2$.
Figure \ref{fig:sim}(b) presents the t-SNE \cite{van2008visualizing} embedding based on the obtained distance matrix between all datasets. The color of each dataset indicates the value of $M$. Indeed, we observe that the datasets are organized according this value.
In other words, our hierarchical geometric-topological analysis provides a metric between datasets that well-captures the similarity in terms of the global structure  of the datasets.

\section{Application to HSI}
\label{sec:application}
In this section we demonstrate the performance of our new geometric-topological framework on Hyper-Spectral Imaging (HSI). 
The structure of hyper-spectral images fits well with the considered hierarchical dataset model -- a dataset here is a single image, and a sample $S_i$ within the dataset is a single square patch. The observations in each patch correspond to the content of the patch at separate spectral bands (see Figure \ref{fig:HSI_hie}).

\begin{figure}[t]
    \centering
    \includegraphics[width=\textwidth]{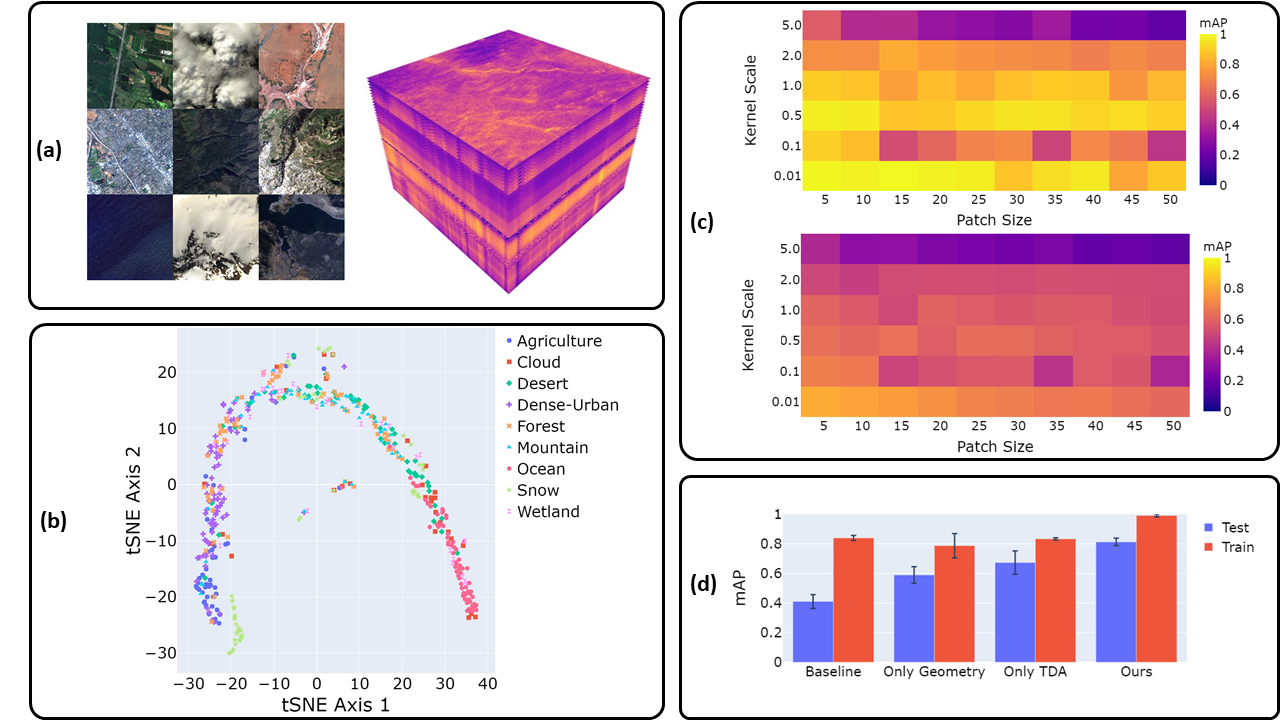}
    \caption{(a) Left: examples of the terrains of hyper-spectral images from different categories (RGB). Right: hyper-spectral image example, stack of $224$ spectral bands.
    (b) t-SNE embedding based on the Wasserstein distance for $H_1$ (holes) between the hyper-spectral images. Each category is denoted by a different color and marker.
    (c) Classification results (mAP score) as a function of two hyper-parameters: kernel scale and patch size. Top: train score. Bottom: test score.
    (d) Numerical ablation study -- classification results (mAP score) evaluating the contribution of each component in our method.} 
    \label{fig:app}
\end{figure}

The HSI database contains images of various terrain patterns, taken from the NASA Jet Propulsion Laboratory's Airborne Visible InfraRed Imaging Spectrometer (AVIRIS) \cite{AVIRISweb}. 
In \cite{ben2019toward} these images were classified into nine categories:
\texttt{agriculture}, \texttt{cloud}, \texttt{desert}, \texttt{dense-urban}, \texttt{forest}, \texttt{mountain}, \texttt{ocean}, \texttt{snow} and \texttt{wetland}. 
The database consists of $486$ hyper-spectral images of size $300 \times 300$ pixels and the spectral radiance is sampled at $224$ contiguous spectral bands ($365$~nm to $2497$~nm). See Figure \ref{fig:app}(a). In terms of our setting, we have $N_D = 486$ datasets, $N = 3600$ samples in each dataset (taking a patch size of $n=5$), and each sample consists of $L = 224$ observations. In this case, each of the observations is a vector in  ${\cal O}_i = \R^{25}$ (corresponding to the patch-size).

The simplicial complex $X$ we use here is a standard triangulation of the 2-dimensional grid of patches.
This way the spatial organization of patches in the image is taken into account in the computation of the persistent homology.

We apply Algorithm \ref{algo:TDA} to the images (datasets) and obtain their pairwise distances.
Figure \ref{fig:app}(b) demonstrates how our new topological distance arranges the images in space. Specifically, we plot the t-SNE embedding \cite{van2008visualizing} of the images (datasets) based on $d_{\mathrm{dataset}}$ for $k=1,p=2$. Each point in the figure represents a single hyper-spectral image, colored by category. Importantly, the category information was not accessible to the (unsupervised) algorithm, and was added to the figure in order to  evaluate the results. 
We observe that most images are grouped by category.
In addition, the embedding also conveys the similarity between different categories, implying that this information is captured by $d_{\mathrm{dataset}}$.
For example, \texttt{agriculture} images (blue points) and \texttt{dense-urban} images  (purple crosses) are embedded in adjacent locations, and indeed they share common patterns (e.g., grass areas). Conversely, \texttt{snow} images form their own separate cluster, as most of the snow instances do not have any common structure with the other categories.

For an objective evaluation of the results, we train an SVM classifier \cite{cortes1995support}.
Prior to computing the SVM, we embed the images (datasets) into a Euclidean space using diffusion maps \cite{Coifman2006}, and  apply the classifier to the embedded images. For diffusion maps, the distance obtained by Algorithm \ref{algo:TDA} $d_{\mathrm{dataset}}$ is used as input, and we generate embedding with $20$ dimensions (see Appendix \ref{sec:appendix_diffusion_maps}).
For the classification, we divide the datasets into a train set and a test set with $10$-fold cross validation; the reported results are the average over all folds.
We use mean Average Precision (mAP) as the evaluation score. 

We compare our results to the results reported in \cite{ben2019toward}, where a deep learning approach was used for the classification of the images. 
To the best of our knowledge, the results in \cite{ben2019toward} are considered the state of the art for the ICONES dataset.
In Table \ref{tab:class_results}, we present the obtained classification results.
In order to make a fair comparison with the reported results in \cite{ben2019toward}, we show the mAP obtained on the train sets. We observe that our method achieves superior results. In addition, we report that our method obtains $0.81$ mAP on the test sets.

In order to test the sensitivity of the proposed algorithm to the choice of hyper-parameters, in Figure \ref{fig:app}(c) we present the train scores (top) and test scores (bottom) as a function of the two key hyper-parameters -- the patch size $n$ and and kernel scale $\epsilon$ (normalized by the median of the distances in the affinity matrix \eqref{eq:kernel}).
The correspondence of the colors between the two figures, as well as the apparent smoothness of the color gradient within each image imply robustness to hyper-parameter tuning. Further, we can optimize the hyper-parameters using the train set without leading to an overfit.

\begin{table}[tb]
\caption{Classification Results (mAP).}
\label{tab:class_results}
\begin{tabular}{|l|l|l|l|l|l|l|l|l|l||l|}
 \hline
  & \texttt{agric.} & \texttt{cloud} & \texttt{desert} & \texttt{dense-urban} & \texttt{forest} & \texttt{mountain} & \texttt{ocean} & \texttt{snow} & \texttt{wetland} & All\\
 \hline \hline
  \cite{ben2019toward}   & 0.48 & 0.66 & 0.5 & 0.86 & 0.57 & 0.64 & 0.83 & 0.57 & 0.23 & 0.59 \\
 \hline
  Ours &   1.0  & 0.95  & 1.0 & 1.0 & 0.99 & 1.0 & 1.0 &  1.0 & 0.82 & 0.98\\
 \hline
\end{tabular}
\end{table}

Next, we perform an ablation study to evaluate the contribution of the geometric and topological analyses separately. In order to do so, 
we consider three variants of the algorithm.
(i) A \emph{`baseline'} solution: here we replace both the geometric and the topological components with the following implementation which was inspired by \cite{ben2019toward}.
We split the $L=224$ spectral bands into $5$ contiguous ranges.
For each range, we apply Principal Component Analysis (PCA) and keep only the principal component in order to reduce the clutter and to get the essence of the spectral information.
Next, the pairwise Euclidean distance between the principal components is considered as the counterpart of $d_{\text{dataset}}$ (the output of Algorithm \ref{algo:TDA}).
% TODO: if we end up with an appendix, then we need to elaborate there because the procedure is not clear. 
%
(ii) \emph{Geometry-based} solution: in Step 1 of Algorithm \ref{algo:TDA}, the weighted simplicial complex is replaced by a weighted graph, taking into account only the weights on the edges $V([i_1,i_2])$. Step 2 is removed, and in Step 3, we use a spectral distance (the $L_2$ distance between the eigenvalues of graphs as in \cite{rajendran2016data}) between the graphs as the output of the algorithm $d_{\text{dataset}}$.
(iii) \emph{Topology-based} solution: in Algorithm \ref{algo:TDA}, the weight function is set to be the cross-correlation between the samples (rather than using alternating diffusion).

We repeat the use of an SVM classifier as described above using the output of each of three variants. Figure \ref{fig:app}(d) shows the results.
First, we observe that the simple baseline based on PCA attains a test score of only $0.41$ mAP.
Second, the addition of the geometric analysis and the topological analysis significantly improves the results.
Third, the combination of the analyses in Algorithm \ref{algo:TDA} gives rise to the best results.

To conclude, rather than relying on the measured values alone, the application of our method to HSI emphasizes associations within the data. This concept is embodied in the proposed hierarchical manner. At the fine scale, graphs based on local spectral associations are constructed. At the coarse scale, simplicial complexes based on global spatial associations are formed. The combination of the structures at the two scales, involving both spectral an spatial information, is shown to be beneficial and gives rise to an informative and useful representation of the images.

%\section{Conclusion}
%\label{sec:conclusion}
%\input{Sections/Conclusion}

\begin{appendix}
\section*{Appendix}
%\section{Code and Data Availability}
%\input{Sections/Code}

\section{Additional Background}
\subsection{Diffusion Maps}
\label{sec:appendix_diffusion_maps}

In Section 6, diffusion maps \cite{Coifman2006} were used for the embedding of the hyper-spectral images (datasets) into a Euclidean space, prior to applying the SVM classifier.
In Algorithm \ref{algo:DM} below, we present the diffusion maps algorithm following the notation in the paper.
Note the use of the distance $d_{\text{dataset}}$ obtained by the proposed method in Step 1. 

Compared to other dimension reduction techniques, the main benefit of diffusion maps is that it embeds the data in a geometrically meaningful Euclidean space.
More concretely, the Euclidean distance between the embedded points (in our notation $\|\tilde{D}_i - \tilde{D}_j\|_2$) approximates the \emph{diffusion distance}, which is an informative notion of distance on the underlying manifold, related to the geodesic distance. Furthermore, as the embedding dimension $d$ increases (it is upper bounded by the number of samples $N_D$), the approximation is more accurate.
For a description of diffusion maps and the diffusion distance in a general context and for more details, see \cite{Coifman2006,talmon2013diffusion}.
% [omer] Why do we used the datasets and d_dataset here? Shouldn't we describe DM in general terms?
\begin{algorithm}[t]
    \SetAlgoLined
        \SetKwInOut{Input}{Input}
        \SetKwInOut{Output}{Output}
        \SetKwInOut{Parameter}{Parameters}
        \Input{Datasets $\{D_i\}_{i=1}^{N_D}$ and a distance metric between datasets $d_{datasets}(D_i,D_j)$.}
        \Output{Diffusion maps embedding $\{\tilde{D}_i\}_{i=1}^N$, $\tilde{D}_i \in \mathbb{R}^d$.}
        \Parameter{Kernel scale: $\epsilon$, embedding dimension: $d$}
        \begin{enumerate}
        \item Compute the affinity matrix $\bm{W} \in \mathbb{R}^{N_D \times N_D}$:
            \[
                W(j_1,j_2) = \exp\left(-\frac{d^2_{\text{dataset}}(D_{j_1}, D_{j_2})}{\epsilon}\right)
            \]
            for all $j_1,j_2=1,\ldots,N_D$
        \item Compute the diagonal matrix: $\bm{Q}=\text{diag}(\bm{W}\bm{1})$
        \item Normalize the affinity matrix: $\widetilde{\bm{W}} = \bm{Q}^{-1} \bm{W} \bm{Q}^{-1}$
        \item Compute the diagonal matrix: $\widetilde{\bm{Q}} = \text{diag}(\widetilde{\bm{W}}\bm{1})$
        \item Compute the diffusion operator: $\bm{K} = \widetilde{\bm{Q}}^{-1} \widetilde{\bm{W}}$
        \item Compute the $d+1$ largest eigenvalues $\lambda_0, \ldots, \lambda_{d}$ and their corresponding (right) eigenvectors $\varphi_0, \ldots, \varphi_{d}$ of $\bm{K}$
        \item Construct a $d$-dimensional embedding for each dataset $D_i, i=1,\ldots, N_D$ by 
        \[
            \tilde{D}_i = (\lambda_1 \varphi_1(i), \ldots, \lambda_{d} \varphi_{d}(i))^T
        \]
        \end{enumerate}
        \caption{Diffusion Maps \cite{Coifman2006}}
        \label{algo:DM}
\end{algorithm}

\subsection{Persistent Homology}
\label{sec:appendix_persistent_homology}
In Section 3.2 we described homology and persistent homology very briefly and in a rather intuitive way. Here, we wish to provide more details.

\begin{figure}
    \centering
    \includegraphics[width=0.9\textwidth]{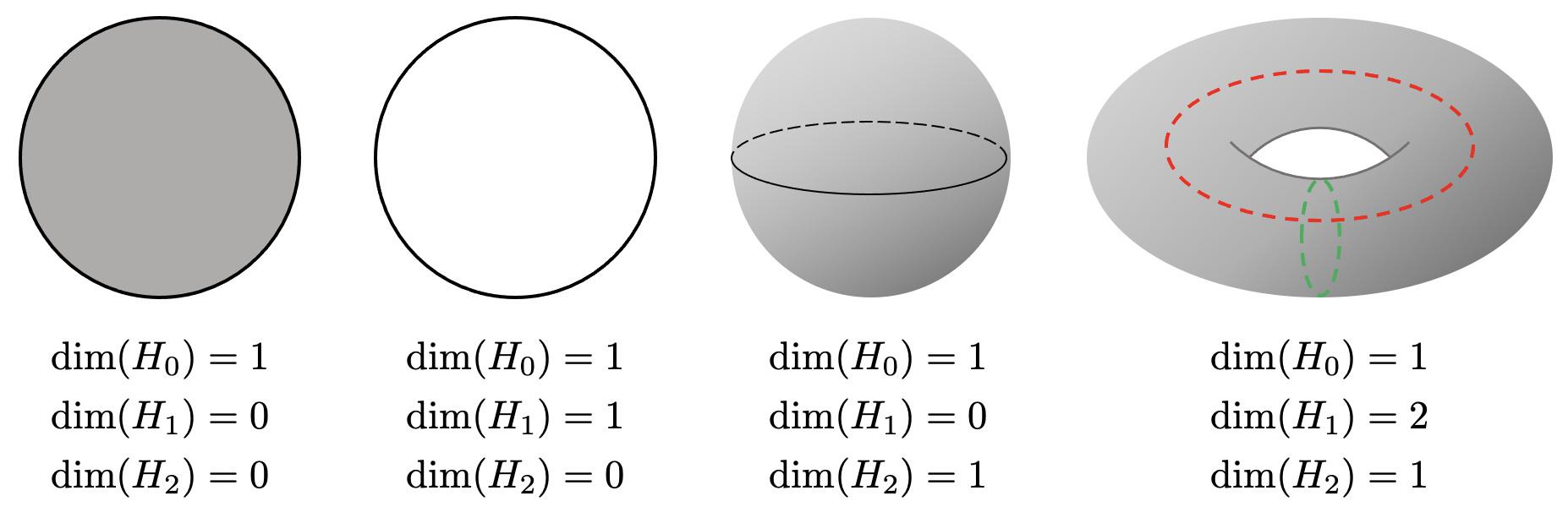}
    \caption{Homology -- examples. We present the dimension of $H_k$ ($k=0,1,2$) in four different shapes: a solid disk, a circle, a sphere and a torus. Note that for the torus, in addition to having a single component ($H_0$) and a single ``air-pocket" ($H_2$) there are two different loops, marked by the dashed lines, that correspond to the basis of $H_1$.}
    \label{fig:homology}
\end{figure}

        \emph{Homology} (cf.~\cite{hatcher2005algebraic,munkres_elements_1984}) is an algebraic structure describing the shape of a topological space. Let $X$ be a topological space (e.g.~a manifold, a simplicial complex, etc.). In its simplest form, the homology of $X$ is a sequence of vector spaces denoted $H_0(X), H_1(X),\ldots$, where each $H_k(X)$ captures the following information. The basis elements of $H_0(X)$ correspond to the connected components of $X$. The basis elements of $H_1(X)$ correspond to closed loops in $X$ that surround a hole (i.e.~not filled in by $X$). The basis elements of $H_2(X)$ correspond to closed 2d surfaces surrounding ``cavities" or ``bubbles" in $X$. Generally, we say that $H_k(X)$ is generated by nontrivial $k$-cycles, which we can be thought of as $k$-dimensional closed surfaces that are not on the boundary of a $(k+1)$ solid. See Figure \ref{fig:homology} for a few examples. 

Homology is relatively simple to define where $X$ is a \emph{simplicial complex}, as discussed in the paper, and where the coefficients used are in the field $\mathbb{Z}_2 = \mathbb{Z}/2$.
Denote by $X^k$ the set of all $k$-simplexes in $X$, and set $F_k = |X^k|$.
We define the \emph{boundary matrix (operator)} $\partial_k$  to be a $F_{k-1}\times F_{k}$ matrix with entries in $\mathbb{Z}_2$, that are set as follows. Let $X^k = \{\sigma_1,\ldots,\sigma_{F_k}\}$ and $X^{k-1} = \{\tau_1,\ldots,\tau_{F_{k-1}}\}$. Then
\[
	(\partial_k)_{i,j} = \begin{cases} 1 & \tau_i \subset \sigma_j,\\ 0 & \text{otherwise}.\end{cases}
\]
In other words, the boundary matrix tells us which $(k-1)$-simplex is on the boundary of which $k$-simplex.
Next, define
\[
	Z_k = \ker(\partial_k),\quad B_k = \mathrm{image}(\partial_{k+1}).
\]
The vector space $Z_k$ contains all combination of $k$-simplexes with no boundary, known as $k$-cycles.
The vector space $B_k$ contains all combination of $k$-simplexes that are the boundary of some $(k+1)$-dimensional structure, known as $k$-boundaries. Finally, the $k$-th homology is the quotient space defined as
\[
	H_k = Z_k / B_k,
\]
i.e.~$H_k$ is generated by all $k$-cycles that are \emph{not} $k$-boundaries. This way, for example, we are able to differentiate between a closed loop that surrounds a hole (= a nontrivial element in $H_1$),  and a closed loop that bounds a 2-dimensional surface (= an element in $B_1$ = a trivial element in $H_1$).
Note that the algebraic formulation of homology suggests that it can be computed using relatively standard matrix diagonalizations steps. This can be seen, for example, in \cite{zomorodian2005computing}.

In addition to describing the shape of a single space, homology can also be used to describe mappings between spaces. Let $X$ and $Y$ be two topological spaces, and let $f:X\to Y$ be a continuous function. We can define a corresponding linear transformation $f_*:H_k(X)\to H_k(Y)$ called the  \emph{induced} map. Intuitively, the function $f_*$ tells us how $k$-cycles in $X$ are mapped to $k$-cycles in $Y$. Such mappings serve as an important ingredient in the definition of \emph{persistent homology} which we discuss next.

\begin{figure}
    \centering
    \includegraphics[width=0.6\textwidth]{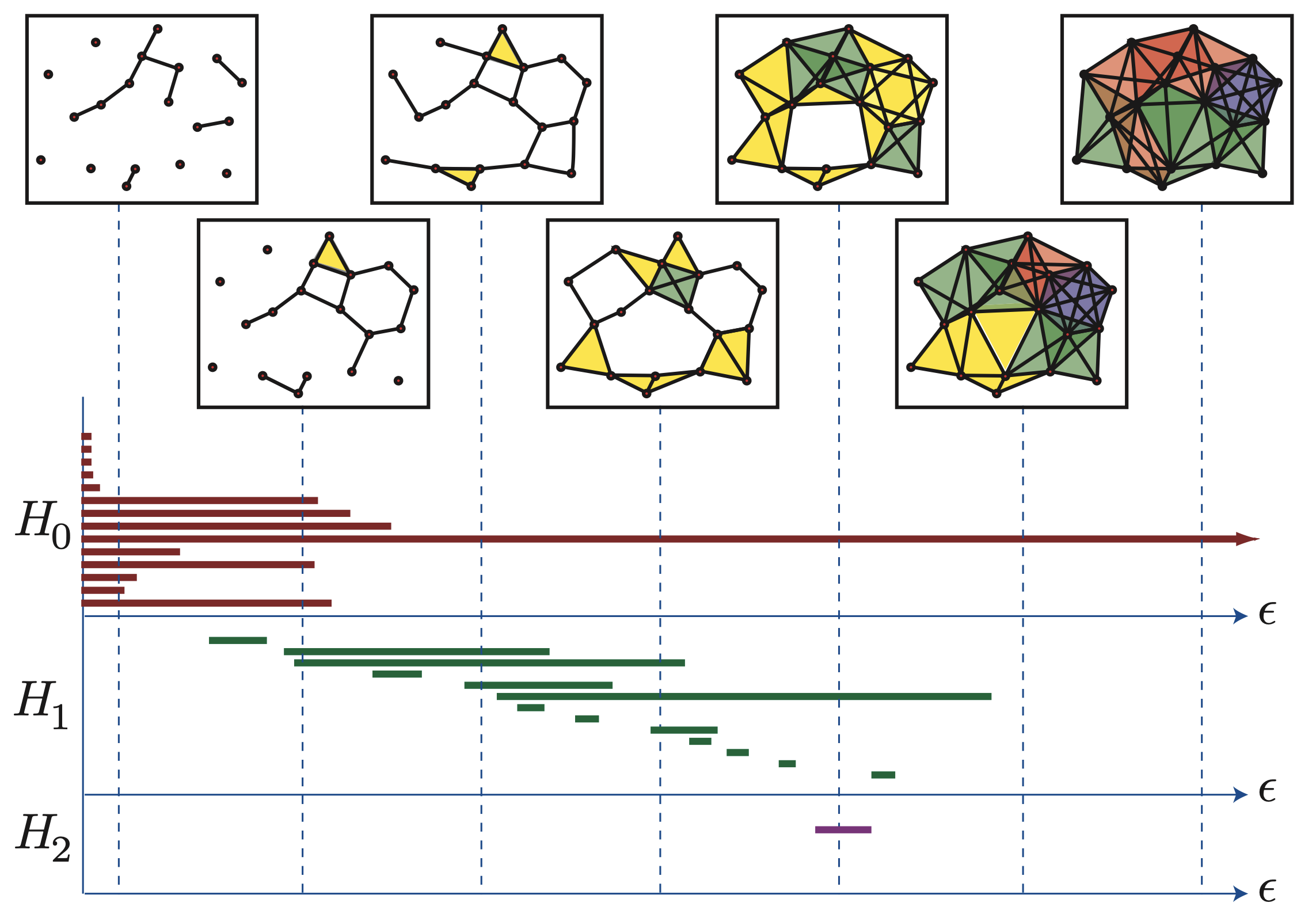}
    \caption{Persistence barcode, taken from \cite{ghrist_barcodes:_2008}. At the top we can see a few snapshots from a filtration of simplicial complexes (parametrized by $\epsilon$). The barcode diagram summarizes the information captured by persistent homology. This information is divided here between $H_0$, $H_1$, and $H_2$. Each $k$-cycle that appears in the filtration is represented by a bar, where the endpoints correspond to its birth and death times. }
    \label{fig:ghrist}
\end{figure}

\emph{Persistent homology} (cf.~\cite{edelsbrunner2008persistent,zomorodian2005computing})  can be viewed as an extension of homology from individual spaces into filtrations of spaces. By a `filtration'  we refer to  a nested sequence  of topological spaces $\{X_t\}_{t\in \R}$, so that $X_s \subset X_t$ for all $s<t$. In this case, for every $s<t$ we can define the inclusion map $i:X_s\hookrightarrow X_t$, which in turn induces the linear map $i_*:H_k(X_s)\to H_k(X_t)$. Using the information provided by all the induced maps ($\forall s,t$), we can track the evolution of every $k$-cycle throughout the filtration, from the moment it first appeared (``born"), to the moment it was terminated (``died").  At the end of this process, the $k$-th persistent homology, denoted $\mathrm{PH}_k$, can be thought of as a collection of $k$-cycles together with their corresponding {lifetime} (i.e. the range of values of $t$ where a cycle exists). While $\mathrm{PH}_k$ is an intricate algebraic structure (a module, to be precise) \cite{crawley-boevey_decomposition_2015}, in most applications, as well as the one presented in the paper, the features used eventually are the $(\text{birth},\text{death})$ values corresponding to each $k$-cycle that appears in a filtration. This information is commonly summarized in a \emph{persistence barcode}  (see Figure \ref{fig:ghrist}), or in a \emph{persistence diagram} (see Figure 1 in the paper).

\section{Hierarchical Structure of Hyper-spectral Images}\label{sec:HSI}

We demonstrate in more detail the setting described in Section 2 in the context of HSI. Here, each dataset $D$ corresponds to a single hyper-spectral image. Note that we allow images to have different sizes.
We split each image $D$ into $N$ patches, constituting the samples $\{S_{i}\}_{i=1}^{N}$.
Recall that each hyper-spectral image $D$ consists of a set of 2-dimensional images corresponding to different spectral bands. The number of spectral bands in the image $D$ is the sample size $L$. 
Splitting each image into patches of size $n\times n$,
the sample target spaces are defined to be ${\cal O}_{i} = \R^{n\times n}$.
In other words, for every dataset (image) we have $L$ observations corresponding to the $L$ spectral bands.
These observations are viewed differently at different patches. The function $f_i$ is thus the expression of each observation (spectral band information) at a particular location (patch). In short, it can be said that $f_i$ encodes spatial information, and the different manifold combinations encode spectral information.
Note, that we have a direct access to neither $\Pi$ nor the functions $f_{i}:\Pi\to {\cal O}_{i}$. Yet, we can think of $\Pi$ as the (high-dimensional) manifold that contains the global information underlying the images from all the different (single) spectral bands.

\end{appendix}

\bibliographystyle{splncs04}
\bibliography{Sections/bib.bib}

\begin{thebibliography}{10}
\providecommand{\url}[1]{\texttt{#1}}
\providecommand{\urlprefix}{URL }
\providecommand{\doi}[1]{https://doi.org/#1}

\bibitem{AVIRISweb}
{NASA} jet propulsion laboratory's airborne visible infrared imaging
  spectrometer ({AVIRIS}). \url{https://aviris.jpl.nasa.gov/}

\bibitem{belkin2003laplacian}
Belkin, M., Niyogi, P.: Laplacian eigenmaps for dimensionality reduction and
  data representation. Neural Comput.  \textbf{15}(6),  1373--1396 (2003)

\bibitem{ben2019toward}
Ben-Ahmed, O., Urruty, T., Richard, N., Fernandez-Maloigne, C.: Toward
  content-based hyperspectral remote sensing image retrieval (cb-hrsir): A
  preliminary study based on spectral sensitivity functions. Remote Sens.
  \textbf{11}(5), ~600 (2019)

\bibitem{berard1994embedding}
B{\'e}rard, P., Besson, G., Gallot, S.: Embedding {R}iemannian manifolds by
  their heat kernel. Geom. Funct. Anal.  \textbf{4}(4),  373--398 (1994)

\bibitem{carlsson_topology_2009}
Carlsson, G.: Topology and data. Bull. Am. Math. Soc.  \textbf{46}(2),
  255--308 (2009)

\bibitem{chang2003hyperspectral}
Chang, C.I.: Hyperspectral imaging: techniques for spectral detection and
  classification, vol.~1. Springer Sci. \& Business Media (2003)

\bibitem{cohen-steiner_lipschitz_2010}
Cohen-Steiner, D., Edelsbrunner, H., Harer, J., Mileyko, Y.: Lipschitz
  functions have {L} p-stable persistence. Found. Comput. Math.
  \textbf{10}(2),  127--139 (jan 2010)

\bibitem{Coifman2006}
Coifman, R.R., Lafon, S.: Diffusion maps. Appl. Comput. Harmon. Anal.
  \textbf{21}(1),  5--30 (2006)

\bibitem{cortes1995support}
Cortes, C., Vapnik, V.: Support-vector networks. Mach. Learn.  \textbf{20}(3),
  273--297 (1995)

\bibitem{crawley-boevey_decomposition_2015}
Crawley-Boevey, W.: Decomposition of pointwise finite-dimensional persistence
  modules. Journal of Algebra and its Applications  \textbf{14}(05),  1550066
  (2015)

\bibitem{dabaghian_topological_2012}
Dabaghian, Y., M{\'e}moli, F., Frank, L., Carlsson, G.: A topological paradigm
  for hippocampal spatial map formation using persistent homology. PLoS Comput.
  Biol.  \textbf{8}(8),  e1002581 (2012)

\bibitem{edelsbrunner2008persistent}
Edelsbrunner, H., Harer, J.: Persistent homology-a survey. Contemporary
  mathematics  \textbf{453},  257--282 (2008)

\bibitem{edgar2002gene}
Edgar, R., Domrachev, M.e.a.: Gene expression omnibus: {NCBI} gene expression
  and hybridization array data repository. Nucleic Acids Res.  \textbf{30}(1),
  207--210 (2002)

\bibitem{ghrist_barcodes:_2008}
Ghrist, R.: Barcodes: the persistent topology of data. Bull. Am. Math. Soc.
  \textbf{45}(1),  61--75 (2008)

\bibitem{giesen2014highly}
Giesen, C., Wang, Hao~AO, e.a.: Highly multiplexed imaging of tumor tissues
  with subcellular resolution by mass cytometry. Nat. Methods  \textbf{11}(4),
  417--422 (2014)

\bibitem{giusti_clique_2015}
Giusti, C., Pastalkova, E., Curto, Carina, e.a.: Clique topology reveals
  intrinsic geometric structure in neural correlations. Proc. Natl. Acad. Sci.
  \textbf{112}(44),  13455--13460 (2015)

\bibitem{hatcher2005algebraic}
Hatcher, A.: Algebraic topology. Cambridge University Press (2005)

\bibitem{jones2008manifold}
Jones, P.W., Maggioni, M., Schul, R.: Manifold parametrizations by
  eigenfunctions of the laplacian and heat kernels. Proc. Natl. Acad. Sci.
  \textbf{105}(6),  1803--1808 (2008)

\bibitem{lamontagne2019oasis}
LaMontagne, P.J., Benzinger, LS, e.a.: Oasis-3: longitudinal neuroimaging,
  clinical, and cognitive dataset for normal aging and alzheimer disease.
  MedRxiv  (2019)

\bibitem{lederman2018learning}
Lederman, R.R., Talmon, R.: Learning the geometry of common latent variables
  using alternating-diffusion. Appl. Comput. Harmon. Anal.  \textbf{44}(3),
  509--536 (2018)

\bibitem{leskovec2014snap}
Leskovec, J., Krevl, A.: {SNAP} datasets: Stanford large network dataset
  collection. \url{http://snap.stanford.edu/data} (2014)

\bibitem{lum_extracting_2013}
Lum, P.Y., Singh, G., Lehman, A., Ishkanov, T., Vejdemo-Johansson, M.,
  Alagappan, M., Carlsson, J., Carlsson, G.: Extracting insights from the shape
  of complex data using topology. Sci. Rep.  \textbf{3}(1), ~1--8 (2013)

\bibitem{van2008visualizing}
Van~der Maaten, L., Hinton, G.: Visualizing data using t-{SNE}. J. Mach. Learn.
  Res.  \textbf{9}(11) (2008)

\bibitem{munkres_elements_1984}
Munkres, J.R.: Elements of algebraic topology, vol.~2. Addison-Wesley Reading
  (1984)

\bibitem{nadler2006diffusion}
Nadler, B., Lafon, S., Coifman, R.R., Kevrekidis, I.G.: Diffusion maps,
  spectral clustering and reaction coordinates of dynamical systems. Appl.
  Comput. Harmon. Anal.  \textbf{21}(1),  113--127 (2006)

\bibitem{naitzat2020topology}
Naitzat, G., Zhitnikov, A., Lim, L.H.: Topology of deep neural networks. J.
  Mach. Learn. Res.  \textbf{21}(184),  1--40 (2020)

\bibitem{rajendran2016data}
Rajendran, K., Kattis, A., Holiday, A., Kondor, R., Kevrekidis, I.G.: Data
  mining when each data point is a network. In: Int. Conf. Patterns of
  Dynamics. pp. 289--317. Springer (2016)

\bibitem{roweis2000nonlinear}
Roweis, S.T., Saul, L.K.: Nonlinear dimensionality reduction by locally linear
  embedding. Sci.  \textbf{290}(5500),  2323--2326 (2000)

\bibitem{shnitzer2019recovering}
Shnitzer, T., Ben-Chen, M., Guibas, L., Talmon, R., Wu, H.T.: Recovering hidden
  components in multimodal data with composite diffusion operators. SIAM J.
  Math. Data Sci.  \textbf{1}(3),  588--616 (2019)

\bibitem{talmon2013diffusion}
Talmon, R., Cohen, I., Gannot, S., Coifman, R.R.: Diffusion maps for signal
  processing: A deeper look at manifold-learning techniques based on kernels
  and graphs. IEEE signal processing magazine  \textbf{30}(4),  75--86 (2013)

\bibitem{talmon2017latent}
Talmon, R., Wu, H.T.: Latent common manifold learning with alternating
  diffusion: analysis and applications. Appl. Comput. Harmon. Anal.
  \textbf{47}(3),  848--892 (2019)

\bibitem{tenenbaum2000global}
Tenenbaum, J.B., De~Silva, V., Langford, J.C.: A global geometric framework for
  nonlinear dimensionality reduction. Sci.  \textbf{290}(5500),  2319--2323
  (2000)

\bibitem{wasserman_topological_2016}
Wasserman, L.: Topological data analysis. Annu. Rev. Stat. Appl.  \textbf{5},
  501--532 (2018)

\bibitem{zomorodian2005computing}
Zomorodian, A., Carlsson, G.: Computing persistent homology. Discrete \&
  Computational Geometry  \textbf{33}(2),  249--274 (2005)

\end{thebibliography}

\end{document}